\documentclass[conference]{IEEEtran}
\IEEEoverridecommandlockouts
\usepackage{algorithm}
\usepackage{algpseudocode}
\usepackage{amsthm}
\usepackage{booktabs}
\usepackage{multirow}
\newtheorem{definition}{Definition}
\usepackage{cite}
\usepackage{amsmath,amssymb,amsfonts}
\usepackage{graphicx}
\usepackage{textcomp}
\usepackage{xcolor}
\usepackage{url}
\usepackage{authblk}
\begin{document}

\title{KG-CF: Knowledge Graph Completion with Context Filtering under the Guidance of Large Language Models
}

\author[$\dagger$]{Zaiyi Zheng}
\author[$\ddagger$]{Yushun Dong}
\author[$\dagger$]{Song Wang}
\author[$\dagger$]{Haochen Liu}
\author[$\text{\textdaggerdbl}$]{Qi Wang}
\author[$\dagger$]{Jundong Li}

\affil[$\dagger$]{\textit{University of Virginia, Charlottesville, USA}, \texttt{\{sjc4fq, sw3wv, sat2pv, jundong\}@virginia.edu}}
\affil[$\ddagger$]{\textit{Florida State University, Tallahassee, USA}, \texttt{yd24f@fsu.edu}}
\affil[$\text{\textdaggerdbl}$]{\textit{Northeastern University, Boston, USA}, \texttt{q.wang@northeastern.edu}}

\maketitle

\begin{abstract}
Large Language Models (LLMs) have shown impressive performance in various tasks, including knowledge graph completion (KGC). However, current studies mostly apply LLMs to classification tasks, like identifying missing triplets, rather than ranking-based tasks, where the model ranks candidate entities based on plausibility. This focus limits the practical use of LLMs in KGC, as real-world applications prioritize highly plausible triplets. Additionally, while graph paths can help infer the existence of missing triplets and improve completion accuracy, they often contain redundant information. To address these issues, we propose KG-CF, a framework tailored for ranking-based KGC tasks. KG-CF leverages LLMs’ reasoning abilities to filter out irrelevant contexts, achieving superior results on real-world datasets. The code and datasets are available at \url{https://anonymous.4open.science/r/KG-CF}.
\end{abstract}


\section{Introduction}
Knowledge Graphs (KGs) have become foundational in numerous real-world applications, including recommendation systems~\cite{bobadilla2013recommender} and knowledge editing~\cite{10.1145/3698590}. Structured as relational data, KGs encode vast factual information using triplets,
where each triplet \texttt{(h, r, t)} specifies a relation \texttt{r} between entities \texttt{h} and \texttt{t} (e.g., \texttt{(Earth, orbits, Sun)}). Despite their utility, KGs are inherently sparse and incomplete, necessitating Knowledge Graph Completion (KGC) to predict missing triplets and thereby enrich the graph~\cite{chen2020knowledge}. Traditional embedding-based methods, such as RotatE~\cite{sun2019rotate}, have shown competitive results in KGC but often lack the ability to integrate external knowledge, including commonsense information not explicitly represented in the graph~\cite{yao2023exploring}. To address this, recent research leverages pretrained language models (PLMs)~\cite{li2022multitask}, with large language models (LLMs) in particular receiving significant attention for their robust reasoning and generalization capabilities~\cite{hao2023reasoning}.

Despite the growing interest in applying LLMs to KGC, current research still encounters significant limitations in practical applications. LLM-based KGC models primarily focus on triplet classification, performing binary evaluations (true or false) on potential missing triplets. However, most real-world KGs are sparse, with an imbalanced distribution of valid and invalid missing triplets. Conversely, traditional approaches (e.g., embedding-based models) emphasize ranking-based tasks, particularly entity prediction, which predicts missing entities for queries in the form \texttt{(?, r, t)} or \texttt{(h, r, ?)}. This task requires models to generate a ranked list of candidate head or tail entities based on relevance and plausibility, forming candidate triplets with the query. Such ranking is crucial in practice, as prioritizing likely missing triplets enhances efficiency and flexibility. However, two intrinsic challenges in LLM-based models hinder their performance on ranking-based tasks:
(1) From the graphs's perspective, existing LLM-based frameworks~\cite{wang2020kepler, chepurova2023better} primarily extract and input graph contextual information (e.g., graph topology, textual descriptions), so-called graph contexts, in textual form to enhance completion. However, in KGC tasks, some extracted graph contexts are irrelevant to the existence of given candidate triplets, introducing substantial redundancy and diverting the LLM's focus from the KGC task.
(2) Sequential generation in LLMs makes them poorly suited for handling numerical values like floats~\cite{jin2024timellm}, complicating the generation of precise plausibility scores for ranking candidate entities or triplets. Standard LLMs produce numbers digit by digit, leading to cumulative sequence errors~\cite{yang2024rethinking}. This limitation also affects generating ranking lists. Moreover, triplet labels used in training are discrete (e.g., true/false), making it challenging to align these labels with continuous score outputs for autoregressive LLMs.

To address these challenges, we introduce KG-CF (\textbf{K}nowledge \textbf{G}raph Completion with \textbf{C}ontext \textbf{F}iltering). In this framework, LLMs are dedicated to filtering out irrelevant contextual information. Specifically, for any triplet \texttt{(h, r, t)} in a knowledge graph $\mathcal{G}$, we sample paths from the head entity \texttt{h} to the tail entity \texttt{t} in $\mathcal{G}$, forming a context set $\mathcal{C}$ to be filtered. The LLM then evaluates $\mathcal{C}$ for relevance to \texttt{(h, r, t)}.
To reduce computational costs, we distill a smaller sequence classifier \( sc \) from the LLM to handle most of the context filtering. This allows us to effectively eliminate irrelevant paths and address the first challenge. Next, we train a smaller PLM, BERT~\cite{devlin2019bert}, on the filtered context set $\mathcal{C^*}$ for path scoring. In the testing phase, we sample the corresponding $\mathcal{C}$ for each triplet and use the highest score from $\mathcal{C}$ as the triplet’s ranking score. By avoiding direct use of the LLM in ranking, we also overcome the second challenge.

Our contributions are summarized in three-fold:
%

\begin{itemize} 
    \item  \textbf{Problem Formulation.} We summarize the challenges related to model design and training data for LLMs in KGC tasks.
    Moreover, we delineate a specific application (context filtering) of LLMs in this scenario.

    \item \textbf{Framework Design.} We propose a principled framework, KG-CF, which successfully leverages the knowledge encoded in the LLMs while still being able to align with the ranking-based tasks and evaluations in KGC.
    
    \item \textbf{Empirical Evaluation.} We conduct empirical evaluations on real-world KG datasets. The experiment results validate the superiority of the proposed model KG-CF compared with other alternatives in KGC tasks. 
\end{itemize}

\begin{figure*}
    \centering
    \vspace{2mm}
    \includegraphics[width=\textwidth]{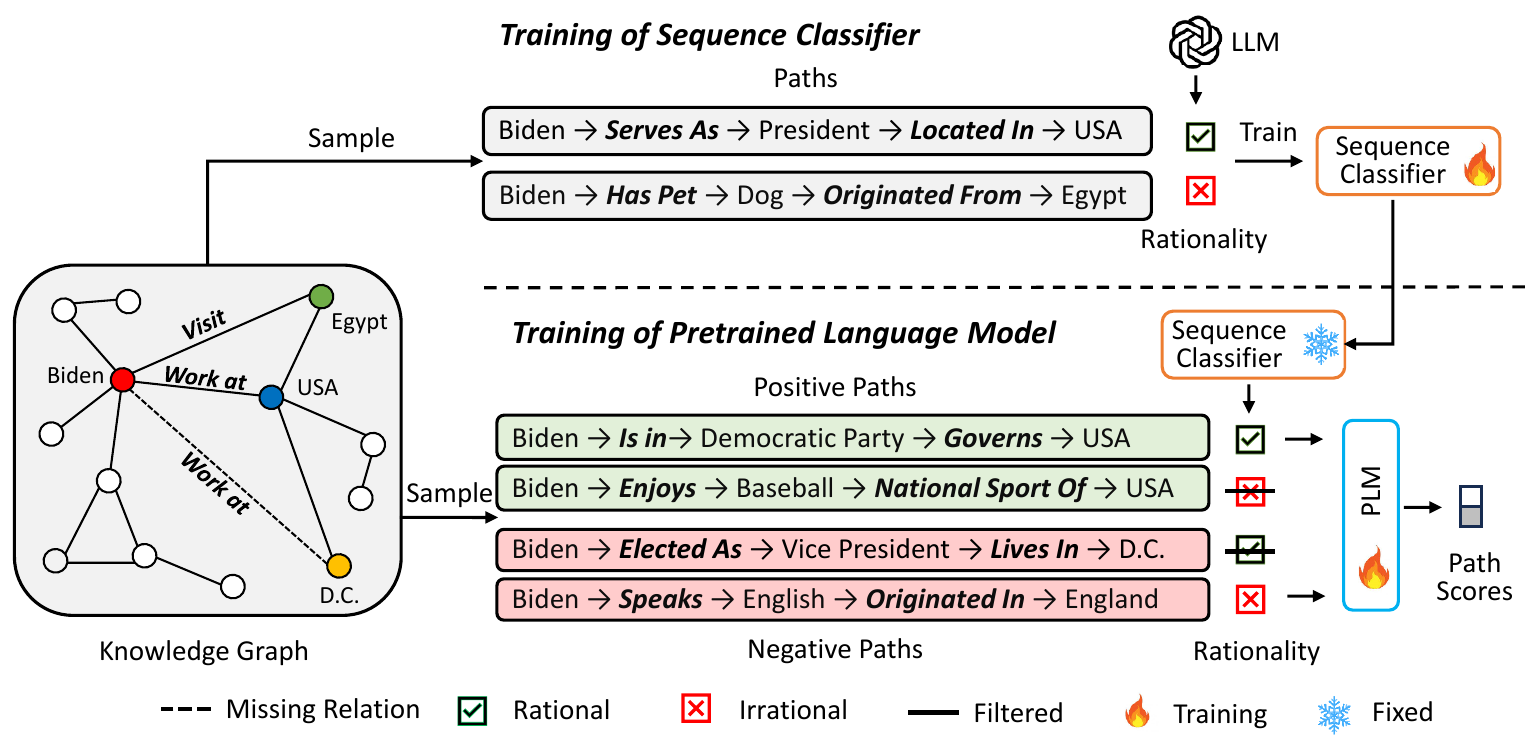}
    \caption{\textbf{The pipeline of KG-CF}. 
    The model operates in three primary steps: 
    1) Sample a small set of paths and use LLMs to generate rationality labels for them. 
    2) Train our sequence classifier on the sampled path set. Then, filter all paths using the sequence classifier, retaining only ``rational'' positive and ``irrational'' negative sample paths. 
    3) Feed all data, including queries, tail nodes, and inference paths, into a PLM for binary classification training. The PLM scorer will output a number between 0 and 1 as the score for the current triplet candidate.
}
    \label{fig:pipeline}
\end{figure*}

\section{Preliminary}

\subsection{Problem Formulation}
We denote the knowledge graph as $\mathcal{G} = \{\mathcal{E}, \mathcal{R}, \mathcal{T}\}$, where $\mathcal{R}$ represents relation types, $\mathcal{E}$ represents entities, and $\mathcal{T}$ includes all triplets in $\mathcal{G}$. A triplet \( t \in \mathcal{T} \) is defined as \( t = (e_h, r, e_t) \), with \( e_h \) as the head entity and \( e_t \) as the tail entity. Our focus is on entity prediction, covering two subtasks: head prediction
and tail prediction~\cite{bordes2013translating}. We provide the definition for tail prediction, with head prediction defined analogously.

\begin{definition}[\textbf{\textit{Tail Entity Prediction}}]
    Given a query $q = (e_h, r_q, ?)$ where $r_q$ is the query relation, we define the completion of $q$ by $e_t$ as:
    \begin{equation}
        c(q, e_t) = q|_{?=e_t} = (e_h, r_q, e_t),
    \end{equation}
    where $c$ denotes the completion function. Firstly, we need to identify the candidate set $\mathcal{C}$ for the tail: \begin{equation} \begin{split} &\mathcal{C} = \{e_i\}_{i=1\rightarrow n}\subseteq \mathcal{E}\setminus \{e_h\}, \\
        &s.t.\ \forall e_t \in \mathcal{C}, c(q, e_t) \notin \mathcal{T},
        \end{split}
    \end{equation}
    where n is a predefined integer. Our objective is to identify a ranking list $A$ of all candidates: 
    \begin{align}
        &\forall i \in [1, n), score(A_i)\geq score(A_{i+1})
    \end{align}
    where $score$ is the scoring function. 
\end{definition}
\noindent \textbf{Example.}
Consider a KG of countries and their capitals. An example query in this graph is presented as follows: 
$$q=(Japan, Capital, ?).$$
We have sampled a series of tail candidates: $\mathcal{C}=\{Paris, Tokyo, Peking, Berlin, Kyoto,$ $London\}$. 
If there already exists a comprehensive KGC model, the ranking list could possibly be: $$A=\{Tokyo, Kyoto, Peking, Paris, London\}.$$



\section{Methodology}

\subsection{Model Overview}

In this section, we present our principled framework, KG-CF, which leverages LLM inference capabilities to train sequence classifiers for context filtering in KGC tasks. Figure~\ref{fig:pipeline} outlines our model pipeline, divided into three key stages: path labeling, sequence classification for filtering, and PLM scoring. To address the exponential growth in path numbers with increasing truncation length, we introduce a sequence classifier to filter paths, leveraging the generalizability of graph topology in knowledge graphs to reduce computational costs.

\subsection{Path Labeling through LLM}

\noindent \textbf{Path Formulation.}
For a query \( q=(e_h, r_q, ?) \) and a potential completion $c(q, e_t)$, we can execute a breadth-first search algorithm on the graph to acquire a straightforward inferential path from \( e_h \) to \( e_t \). Each trajectory $T$ is formulated as a list of triplets $\{t_i\}_{i=0\rightarrow n}$ that starts from $e_h$ and ends at a potential tail entity $e_t$:
$$T=( (e_h, r_0, e_1), (e_1, r_1, e_2), ...., (e_{n}, r_{n}, e_t)).$$
We define an inference path $P$ as the concatation of a trajectory $T_q$ along with the completion $c(q, e_t) = (e_h, r_q, e_t)$: 
\begin{equation}
    P= ((e_h, r_q, e_t), T).
\end{equation}

\noindent \textbf{LLM Inference.} \label{llminfer}
So far, we have formalized the objects that need to be filtered. Subsequently, we transform the paths into character sequences to adapt the inference paths to the input of LLMs. 
Therefore, we obtain labels for all the paths associated with $c(q, e_t)$:
\begin{equation}
    \mathcal{Y}_{c(q, e_t)} = LLM(instruction\oplus f(\mathcal{P}_{c(q, e_t)})),
    \label{equ:llminfer}
\end{equation}
where $\oplus$ denotes the concatenation operation, $\mathcal{P}_{c(q, e_t)}$ contains all the possible paths related to $c(q, e_t)$ and $f$ transform the paths into texts. The result $\mathcal{Y}_{c(q, e_t)}$ contains labels for paths in $\mathcal{P}_{c(q, e_t)}$ while each label is in $\{0, 1\}$. Based on this operation, we construct a dataset $\mathcal{D}_{sc}$ for the sequence classifier training, and we introduce the details in the next section. The detailed process is presented in Algorithm~\ref{alg:Dsc}. 

Note that inverse relationships are allowed (suitable for head prediction) and all triplets in the path are represented in the standard forward order. For example, triplet $(Lakers, inv(plays\ for), Lebron\ James)$ will be interpreted as \textit{``Lebron\ James plays for Lakers''}, where $inv()$ represents the function of inversing. 


\begin{algorithm}[!ht] 
\caption{Dataset for Sequence Classifer}
\label{alg:Dsc}
\begin{algorithmic}[1]
\Require KG $\mathcal{G}=(\mathcal{E}, \mathcal{R}, \mathcal{T})$, Maximum path length $m$ and path numbers per relation $n$.  
\Ensure Dataset $\mathcal{D}_{sc}$ for Sequence Classifer.
\State $\mathcal{D}_{sc} \gets \emptyset$
\ForAll{$r \in \mathcal{R}$}
    \State $r_{count}\gets 0$
\EndFor
\ForAll{triples $t \in T$}
    \State $e_h, r, e_t \gets t$
    \If{$r_{count} > n$}
        \State continue
    \EndIf
    \State $\mathcal{P}$ $\gets$ All simple paths from $e_h$ to $e_t\in \mathcal{T}\setminus \{t\}$ with path length up to $m$
    \State $\mathcal{L} \gets \text{Label each path using LLM}$
    \State $\mathcal{D}_{sc} \gets \mathcal{D}_{sc} \cup \{(\mathcal{P}[i], \mathcal{L}[i]) \mid 0\leq i \leq |\mathcal{P}|\}$
    \State $r_{count} \gets r_{count} + 1$
\EndFor
\State \Return $\mathcal{D}_{sc} $
\end{algorithmic}
\end{algorithm}

\begin{algorithm}[!ht] 
\caption{Dataset for PLM}
\label{alg:Dbert}
\begin{algorithmic}[1]
\Require KG $\mathcal{G}=(\mathcal{E}, \mathcal{R}, \mathcal{T})$, Number of negative instances $neg\_num$, Threshold $th$, Maximum path length $m$, Sequence Classifier $sc$.
\Ensure Dataset $\mathcal{D}_{PLM}$ for PLM training.
\State $\mathcal{D}_{PLM} \gets \emptyset$
\ForAll{triples $t \in \mathcal{T}$}
    \State $e_h, r, e_t \gets t$
    \State $\mathcal{P}_{pos} \gets$ All simple paths from $e_h$ to $e_t\in \mathcal{T}\setminus \{t\}$  with up to $m$
    \State $\mathcal{P}_{pos} \gets \{p|p\in\mathcal{P}_{pos}\land sc(p)>th\}$
    \State $\mathcal{D}_{pos}\gets \{(p, true) \mid p \in \mathcal{P}_{pos}\}$
    \State $\mathcal{D}_{PLM} \gets \mathcal{D}_{PLM} \cup \mathcal{D}_{pos}$
    \For{$i \gets 1$ \textbf{to} $neg\_num$}
        \State Pick an $e \in \mathcal{E}\setminus\{e_h\} \ s.t.\ (e_h, r, e)\notin \mathcal{T}$ 
        \State $\mathcal{P}_{neg}\gets$ All simple paths from $e_h$ to $e_t\in \mathcal{T}$ with path length up to $max\_hops$
        \State $\mathcal{P}_{neg} \gets \{p|p\in\mathcal{P}_{neg}\land sc(p)<th\}$
        \State $\mathcal{D}_{neg}\gets \{(p, false) \mid p \in \mathcal{P}_{neg}\}$
        \State $\mathcal{D}_{PLM} \gets \mathcal{D}_{PLM} \cup \mathcal{D}_{neg}$
    \EndFor
\EndFor
\State \Return $\mathcal{D}_{PLM}$
\end{algorithmic}
\end{algorithm}

\subsection{Sequence Classifier}
We employ an LSTM~\cite{hochreiter1997long} model as the sequence classifier \(M_{sc}: \mathcal{P} \rightarrow \{0, 1\}\) to implements functionality similar to LLM in Equation~(\ref{equ:llminfer}). 
Considering a path $P=((e_h, r_q, e_t), ((e_h, r_0, e_1), ..., (e_{n-1}, r_{n-1}, e_t)))$, we have:
\begin{equation}
    \begin{aligned}
    \boldsymbol{h_0} &= R(0, \boldsymbol{e_h}\oplus\boldsymbol{r_0}\oplus\boldsymbol{e_1}\oplus\boldsymbol{r_q}),\\
    \boldsymbol{h_i} &= R(\boldsymbol{h_{i-1}}, \boldsymbol{e_i}\oplus\boldsymbol{r_i}\oplus\boldsymbol{e_{i+1}}\oplus\boldsymbol{r_q}), i\leq n-1,\\
    \hat{y} &= \sigma(fc(h_{n-1})),
\end{aligned}
\end{equation}
where $R$ denotes the LSTM model, $\hat{y}$ is the prediction by applying classifier layer $fc$ and Sigmoid function $\sigma$ to the last hidden state $h_{n-1}$. 
In particular, we use the sequence classifier to filter and construct the dataset $\mathcal{D}_{plm}$ for PLM model training in Sec.~\ref{subsec:bertscoring}. The detailed process is described in Algorithm~\ref{alg:Dbert}.

\noindent \textbf{Optimization.} We use the cross-entropy loss to train the sequence classifier model:
\begin{equation}\label{equ:scloss}
    \mathcal{L} = \sum_{i=1}^{N} \left[ y_i \log(\hat{y}_i)+ (1 - y_i) \log(1 - \hat{y}_i) \right].
\end{equation}
Here, \(N\) is the number of samples, \(y_i\) represents the true label of the \(i\)-th sample (with a value of 0 or 1), and \(\hat{y}_i\) denotes the predicted probability of the \(i\)-th sample being in class 1. The label $y$ is generated by LLMs. 




\subsection{PLM Scoring} \label{subsec:bertscoring}
In this section, we demonstrate the scoring and training process of our PLM scorer. Considering a path $P= (c(q, e_t), T)$, we compute the text representation and its score as follows:
\begin{align}
    P_{text} = text(c(q, e_t)) \otimes text(T),\\
    score(P) = \hat{y}_P = \sigma(PLM(P_{text})),
\end{align}
where $text(\cdot)$ stands for the textualize function, $\otimes$ denotes concatenating and independently annotating two segments of text, and $\hat{y}_P$ is the score of the path $P$ by applying the sigmoid function $\sigma(\cdot)$ on the outputs of the PLM model.
We utilize the same loss function as Eq.~(\ref{equ:scloss}) for PLM training, while the label represent the existence of triplets in KG.

\noindent \textbf{Scoring and Ranking.}
To provide a basis for entity ranking, inspired by BERTRL~\cite{zha2021inductive}, we represent the confidence score of each completion $c(q, e_t)$ using the highest path score corresponding to it:
\begin{align}
    &score(e_t) = max\{\hat{y}_P|P\in \mathcal{P}_{c(q, e_t)}\}
\end{align}
A special case occurs when $\mathcal{P}_{c(q, e_t)} =\emptyset$. In this scenario, we manually assign the lowest score to the completion. 

\begin{table}
\centering
\caption{Performances on \textbf{transductive} entity prediction of traditional (top) and PLM-based approaches (bottom). Results are in percentage with the best ones shown in \textbf{Bold}. } 
\begin{tabular}{@{}lcccccc@{}}
\toprule
\multirow{2}{*}{\textbf{Datasets}}  & \multicolumn{2}{c}{\textbf{WN18RR}}  & \multicolumn{2}{c}{\textbf{FB15K-237}} & \multicolumn{2}{c}{\textbf{NELL-995}} \\ 
\cmidrule(r){2-3} \cmidrule(l){4-5} \cmidrule(l){6-7}
& \textbf{Hits@1} & \textbf{MRR} & \textbf{Hits@1} & \textbf{MRR} & \textbf{Hits@1} & \textbf{MRR}\\ 
\midrule
\textbf{RuleN} & 64.6 & 67.1 & 60.2 & 67.5 & 63.6 & 73.7\\
\textbf{GRAIL} & 64.4 & 67.6 & 49.4 & 59.7 & 61.5 & 72.7\\
\textbf{MINERVA}  & 63.2 & 65.6 & 53.4 & 57.2 & 55.3 & 59.2\\
\textbf{TuckER}  & 60.0 & 64.6 & 61.5 & 68.2 & 72.9 & 80.0\\
\midrule
\textbf{BERTRL} & 66.3 & 68.7 & 61.9 & 69.6& 68.6 & 78.2\\
\textbf{KG-CF} & \textbf{67.5} & \textbf{70.3} & \textbf{62.3} & \textbf{70.9} & \textbf{73.1} & \textbf{82.0}\\
\bottomrule
\end{tabular}
\clearpage
\label{tab:mainresult1}

\end{table}

\begin{table}
\centering
\caption{Performances on \textbf{inductive} entity prediction. 
} 
\begin{tabular}{@{}lcccccc@{}}
\toprule
\multirow{2}{*}{\textbf{Datasets}}  & \multicolumn{2}{c}{\textbf{WN18RR}}  & \multicolumn{2}{c}{\textbf{FB15K-237}} & \multicolumn{2}{c}{\textbf{NELL-995}} \\ 
\cmidrule(r){2-3} \cmidrule(l){4-5} \cmidrule(l){6-7}
& \textbf{Hits@1} & \textbf{MRR} & \textbf{Hits@1} & \textbf{MRR} & \textbf{Hits@1} & \textbf{MRR}\\ 
\midrule
\textbf{RuleN} & 74.6 & 78.2 & 41.5 & 46.3 & 63.8 & 71.1\\
\textbf{GRAIL} & 76.9 & 79.9 & 39.0 & 46.9 & 55.4 & 67.5\\
\midrule
\textbf{KG-BERT} & 43.6 & 57.4 & 34.1 & 50.0 & 24.4 & 41.9\\
\textbf{BERTRL} & 75.3 & 79.5 & \textbf{54.1} & \textbf{60.6} & 71.7 & 81.0\\
\textbf{KG-CF} & \textbf{78.5} & \textbf{80.9} & 51.2 & 58.3 & \textbf{79.5} & \textbf{86.6}\\
\bottomrule
\end{tabular}
\clearpage
\label{tab:mainresult2}

\end{table}

\section{Empirical Evaluation}
In this section, we will answer the following four questions through experiments:
(1) How well can KG-CF perform in knowledge graph completion tasks?
%
%
(2) How do different filtering choices contribute to the overall performance of KG-CF?
(3) How does the maximum path length affect the accuracy of the completion?



\subsection{Experimental Settings}


\noindent\textbf{Datasets.}
\label{sec:dataset}  
Three widely utilized real-world knowledge graphs: NELL-995~\cite{xiong-etal-2017-deeppath}, FB15K-237~\cite{bordes2013translating}, and WN18RR~\cite{shang2018endtoend}. NELL-995 and FB15K-237 are relation extraction datasets sourced from web text. WN18RR is derived from WordNet with refined relations. To streamline training, we sample a subset from each source dataset for evaluation.

\noindent\textbf{Baselines.}  
We incorporate methods from prior research as baselines. Among them, RuleN~\cite{meilicke2018fine} (rule-based) and GRAIL~\cite{teru2020inductive} (GNN-based) support both inductive and transductive scenarios, while MINERVA~\cite{das2018walk} (reinforcement learning) and TuckER~\cite{Balazevic_2019} (embedding-based) are limited to transductive settings. Additionally, we include KG-BERT~\cite{yao2019kgbert} and BERT-RL~\cite{zha2021inductive}, both leveraging pretrained language models.


\subsection{Evaluation Method}
\textcolor{black}{In both transductive and inductive scenarios, we separately evaluate our approach on two subtasks: tail prediction and head prediction. The average metrics of two scenarios are shown as the final results.}
Following GRAIL~\cite{teru2020inductive} and BERTRL~\cite{zha2021inductive}, we randomly select another 49 tail entities $\{t_i\}_{i=1\rightarrow49}$ for each test triplet $(h_{test}, r_{test}, t_{test})$ and form a candidate set $T_{test} = \{t_{test}\}\cup\{t_i\}_{i=1\rightarrow49}$. Despite $t_{test}$, we make sure that for any other $t\in T$, $(h_{test}, r_{test}, t) \notin \mathcal{G}$. By the end, we rank $t_{test}$ based on scores and compute metrics. 




\subsection{Main Results (Question 1)}


In this subsection, we assess our KG-CF framework on three knowledge graphs across transductive (Table I) and inductive scenarios (Table II), leading to these insights: (1) KG-CF surpasses most baselines across datasets and scenarios, highlighting the effectiveness of using LLMs and sequence classifiers to refine graph context. (2) KG-CF demonstrates greater stability in transductive scenarios compared to inductive ones. (3) Our method achieves the most significant improvements on NELL-995, which, unlike FB15K-237, includes richer textual descriptions of entities (e.g., “person Mexico Ryan Whitney” rather than “Ryan Whitney”). This detail allows the LLM to better handle rare nouns, enhancing its precision.

\begin{figure*}
    \centering
    \includegraphics[width=\textwidth]{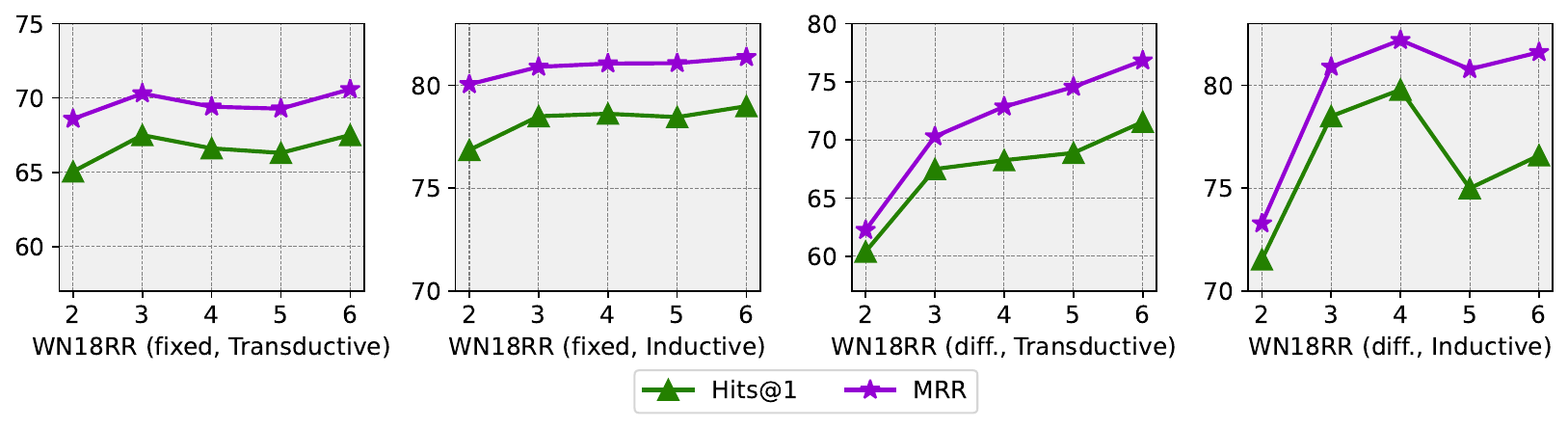}
    \caption{ \textbf{Path length scalability study. }The horizontal axis represents the maximum path length, and the vertical axis represents the metric values.
    }
    \label{fig:scalability}
\end{figure*}

\begin{figure}[!ht]
    \centering
    \includegraphics[width=0.99 \columnwidth]{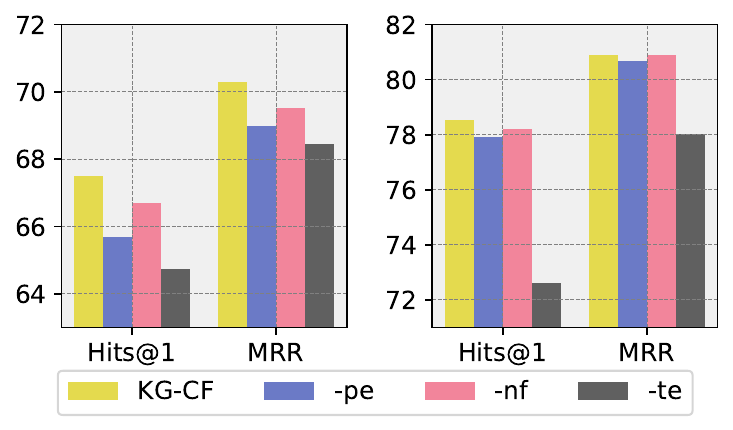}
    \vspace{-9mm}
    \caption{Transductive (left) and inductive (right) performance comparison between KG-CF, KG-CF-pf, KG-CF-nf, and KG-CF-te. Here, -pf, -nf, -te represent positive path filtering, negative path filtering, and trajectory entities being removed, respectively.}
    \vspace{-3mm}
    \label{fig:ablation}
\end{figure}

\subsection{Ablation Study (Question 2)}
We conducted an ablation study on the WN18RR dataset (both transductive and inductive) where three components are removed separately: positive path filtering ($-pf$), negative path filtering ($-nf$), and trajectory entities in the paths ($-te$, i.e., relation only). We present the results in Figure{~\ref{fig:ablation}}.

\noindent\textbf{Positive Path Filtering.} 
Under this setting, we assume that all paths from \( e_h \) to \( e_t \) in the positive triplet \( (e_h, r_q, e_t) \) conform to standard reasoning logic, thus preserved during the data filtering phase. 
The results showed a slight decline compared to the original model, indicating that our sequence classifier can enhance the rationality of positive paths.

\noindent\textbf{Negative Path Filtering.}
In this setting, we assume that for a negative triplet \( (e_h, r_q, e_t) \notin \mathcal{T} \), all paths from \( e_h \) to \( e_t \) fail to confirm \( r_q \)'s existence (though, given KG incompleteness, we consider this assumption incorrect). This ablation led to a slight performance drop, suggesting the sequence classifier effectively filters out false negatives caused by KG incompleteness. In contrast, the performance drop is notably smaller than in -pf, indicating that irrelevant contextual information from existing KG triplets has a more substantial negative impact on performance than missing triplets.

\noindent\textbf{Trajectory Entities.} 
In this ablation, we replaced all entity names in path trajectories with anonymous labels (e.g., "entity1") but retained relation information to test the filtering of our model based solely on topology. This aims to detect data leakage, assessing whether internal knowledge in language models provides an unfair advantage. The significant performance drop confirms this issue, supporting our finding in Section 4.4 that textual descriptions strongly influence LLM filtering. Additionally, with fewer paths in this set-up, performance declines due to reduced dataset size. 
In the inductive scenario, these findings hold in all ablation types, confirming the robustness and effectiveness of KG-CF.

\subsection{Path Length Scalability Study (Question 3)}

Intuitively, providing a knowledge graph completion model with richer contextual information can boost confidence and accuracy in training and prediction. However, the exponential cost of graph sampling at larger scales limits the practical amount of context that can be used. Here, we evaluate the impact of scalability on model performance in two settings.

\noindent\textbf{Fixed Setting.}  
We create training datasets and train models with varying maximum path lengths, but fix the maximum path length at 3 during testing for all models.

\noindent\textbf{Diff. Setting.}  
Training follows the same approach as in the fixed setting, but each model is tested with the maximum path length used during its training.

We run experiments on the WN18RR dataset, showing results for both settings in Figure~\ref{fig:scalability}. Key observations are as follows:
(1) Extending path length generally improves model performance.
(2) Increasing path length from two to three significantly boosts performance in both transductive and inductive scenarios.
However, performance differences for path lengths of three or more are minimal, likely because shorter paths capture more relevant information.
(3) In some cases, training on longer paths may reduce the model’s ability to reason over shorter paths.

\section{Related Work}

\subsection{Knowledge Graph Completion (KGC)}

Existing KGC methods generally fall into three categories:
(1) \textit{Embedding-based Methods:} These approaches map entities and relations into an embedding space, with notable methods like TransE~\cite{bordes2013translating} and DistMult~\cite{yang2015embedding}. Among them, RotatE is widely recognized for its geometric approach to relational semantics.
(2) \textit{Causality-based Methods: }This category~\cite{das2020probabilisticcasebasedreasoningopenworld} focuses on identifying and utilizing causal relationships within knowledge graphs.
(3) \textit{PLM-based Methods: }Initiated by KG-BERT~\cite{yao2019kgbert}, which leverages BERT’s inherent knowledge, this approach has evolved with models like BERTRL~\cite{zha2021inductive} that incorporate reasoning paths. Advanced techniques include prompt engineering, LoRA adapters~\cite{hu2021lora}, and innovations in soft prompts~\cite{qin2021learningaskqueryinglms} for enhanced training and evaluation.

\subsection{Reasoning with Large Language Models}
Currently, there are primarily two strategies~\cite{qiao2023reasoning} for leveraging LLMs in reasoning tasks:
(1) \textit{Strategy Enhanced Reasoning. }
These methods focus on refining the reasoning capabilities and strategies of LLMs. Since LLMs excel in interpreting and following explicit instructions~\cite{liu2023prompting}, prompt engineering~\cite{sahoo2024systematic} has been widely used to directly improve model performance~\cite{wei2023chainofthought}. Additionally, iterative methods~\cite{zelikman2022star} enhance reasoning processes, and external reasoning tools (e.g., code interpreters)~\cite{lyu2023faithful} are employed to support LLMs.
(2) \textit{Knowledge Enhanced Reasoning.  }
Knowledge is crucial for AI reasoning systems~\cite{pan2024unifying}. Some studies~\cite{liu2022generated} focus on extracting information embedded within LLMs, while others incorporate external data sources~\cite{yang2022logicsolver}.

\section{Limitaions}

This work mainly focuses on the problem of utilizing LLM's reasoning ability on KGC. 
We note that we only deploy simple reasoning paths as the graph context, which is not essential for evaluation. Therefore, new context type selection (e.g. ego-graph) can be a future direction that is worthwhile to explore.

\section{Conclusion \& Future Works}
This paper presents KG-CF, a knowledge graph completion method that enhances pretrained language models (PLMs) through LLM-guided context filtering. We distill a sequence classifier from an LLM to assess reasoning path validity, enabling high-quality KG context selection for training the BERT scorer. Experiments show KG-CF achieves strong performance across datasets and scenarios. Our approach efficiently applies autoregressive LLMs to entity ranking. We leave the incorporation of varying graph context types to future works. 
\section*{Acknowledgement}
This work is supported in part by the National Science Foundation under grants (IIS-2006844, IIS-2144209, IIS-2223769, CNS-2154962, BCS-2228534, and CMMI-2411248), the Commonwealth Cyber Initiative Awards under grants (VV-1Q24-011, VV-1Q25-004), and the research gift funding from Netflix and Snap.

\bibliographystyle{IEEEtran}
\bibliography{IEEEabrv, custom}

\end{document}